\documentclass[3p,times,procedia]{elsarticle}

\usepackage{ecrc}
\usepackage{amsmath}
\usepackage{todonotes}

\volume{00}

\firstpage{1}

\journalname{Procedia Computer Science}

\runauth{M.D. Danu et al.}

\jid{procs}

\jnltitlelogo{Procedia Computer Science}

\CopyrightLine{2023}{The Authors. Published by Elsevier B.V.\newline Selection and/or peer-review under responsibility of ITQM 2023}

\usepackage{amssymb}
\usepackage{enumitem}

\usepackage[figuresright]{rotating}

\begin{document}

\begin{frontmatter}

\dochead{Information Technology and Quantitative Management (ITQM 2023)}

\title{Generation of Radiology Findings in Chest X-Ray by Leveraging Collaborative Knowledge}

\author[a,b,*]{Manuela Daniela Danu}
\author[a]{George Marica}
\author[c]{Sanjeev Kumar Karn}
\author[c]{Bogdan Georgescu}
\author[c]{Awais Mansoor}
\author[d]{Florin Ghesu}
\author[a,b]{Lucian Mihai Itu}
\author[a,b]{Constantin Suciu}
\author[c]{Sasa Grbic}
\author[c]{Oladimeji Farri}
\author[c]{Dorin Comaniciu}

\address[a]{Advanta, Siemens SRL, 15 Noiembrie Bvd, 500097 Brasov, Romania}
\address[b]{Automation and Information Technology, Transilvania University of Brasov, Mihai Viteazu nr. 5, 5000174 Brasov, Romania}
\address[c]{Digital Technology and Innovation, Siemens Healthineers, 755 College Rd E, 08540 Princeton, NJ, United States}
\address[d]{Digital Technology and Innovation, Siemens Healthineers, Henkestr. 127, 91052 Erlangen, Germany}

\begin{abstract}
Among all the sub-sections in a typical radiology report, the \emph{Clinical Indications}, \emph{Findings}, and \emph{Impression} often reflect important details about the health status of a patient. The information included in \emph{Impression} is also often covered in \emph{Findings}. While \emph{Findings} and \emph{Impression} can be deduced by inspecting the image, \emph{Clinical Indications} often require additional context. The cognitive task of interpreting medical images remains the most critical and often time-consuming step in the radiology workflow. Instead of generating an end-to-end radiology report, in this paper, we focus on generating the \emph{Findings} from automated interpretation of medical images, specifically chest X-rays (CXRs). Thus, this work focuses on reducing the workload of radiologists who spend most of their time either writing or narrating the \emph{Findings}. Unlike past research, which addresses radiology report generation as a single-step image captioning task, we have further taken into consideration the complexity of interpreting CXR images and propose a two-step approach: (a) detecting the regions with abnormalities in the image, and (b) generating relevant text for regions with abnormalities by employing a generative large language model (LLM). This two-step approach introduces a layer of interpretability and aligns the framework with the systematic reasoning that radiologists use when reviewing a CXR.


\end{abstract}


\end{frontmatter}

\correspondingauthor[*]{Corresponding author.}
\email{manuela.danu@siemens.com}


\section{Introduction}
\label{introduction}
Recent efforts to jointly represent images and text for addressing tasks that require reasoning over multi-modal input have led to the development of a wide range of vision-language (VL) model architectures following the popular transfer learning paradigm of pre-training and fine-tuning \cite{DBLP:journals/corr/abs-1908-06066}, \cite{DBLP:journals/corr/abs-1909-11059}. While the current VL pre-trained models have been utilized for various downstream tasks such as visual question answering \cite{anderson2018bottom}, \cite{lei2018tvqa}, and referring expression comprehension \cite{yu2018mattnet}, their impact on VL tasks that are specific to medical images and radiology workflow has not been thoroughly investigated. Our research aims to apply collaborative knowledge transfer from computer vision and large language models to generate image-to-text, as a version of the multi-modal transfer learning framework.

\begin{figure}[t]
\centerline{\includegraphics[scale=0.25]{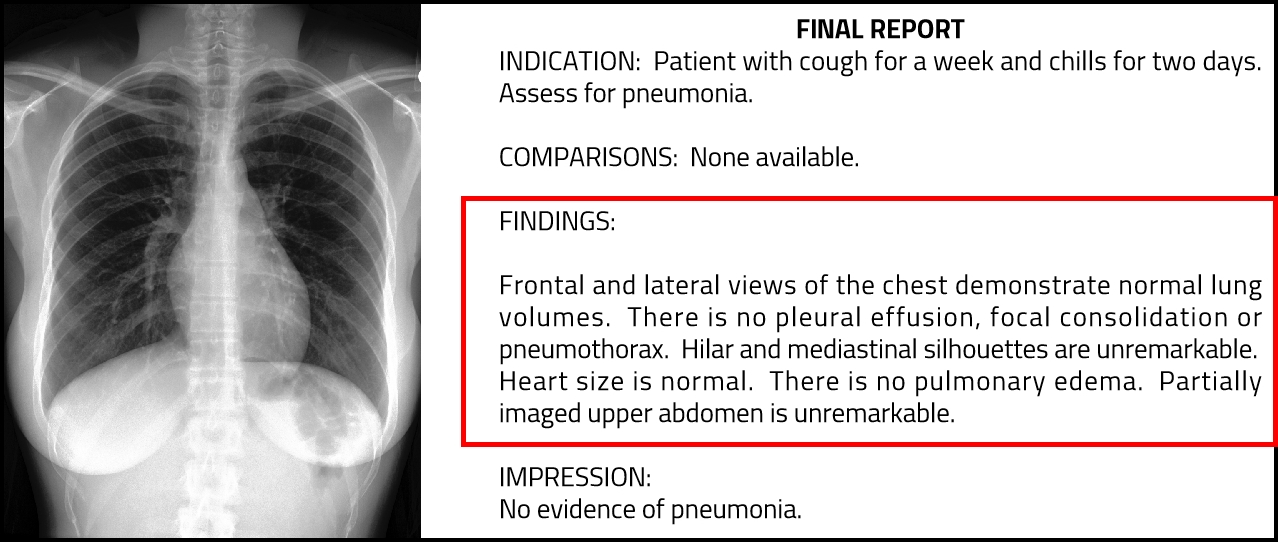}}
\label{fig:mimic-example}
\caption{Example of chest X-ray image and radiology report, extracted from \emph{MIMIC-CXR} dataset.}
\end{figure}

Among all the subsections in a typical radiology report, the {\it Clinical Indications}, {\it Findings}, and {\it Impression} often reflect critical details that should be communicated between radiologists and ordering physicians (Figure \ref{fig:mimic-example}). Specifically, normal anatomical structures and existing abnormalities noticed in the medical images during the radiologist's review are documented as {\it Findings}. Previous studies have revealed the potential for missed abnormalities due to perceptual errors, which could lead to incorrect or delayed diagnoses \cite{degnan2019perceptual}, \cite{waite2019analysis}, and adverse repercussions for the patient. This work aims to reduce these perceptual errors by investigating the adaptation of a state-of-the-art VL model architecture to generate {\it Findings} from CXR images, and could, in turn, contribute to an automated radiology reporting system.


Although there have been previous works on generating part of or the entire radiology report as a single-step image captioning task \cite{chen2020generating}, we have further considered the complexity of reading CXR images and proposed a two-step approach: (a) detecting the regions with abnormalities in the image, using a technique that combines contrastive learning \cite{hadsell2006dimensionality} and online feature clustering \cite{caron2020unsupervised} to achieve self-supervised learning \cite{ghesu2022self}, and (b) generating text relevant to each of the regions with abnormalities by leveraging RadBloomz \cite{karn2023shs}, a fine-tuned version of the Bloomz-7b1 \cite{muennighoff2022crosslingual} generative LLM. This two step approach (Figure 2) introduces a layer of interpretability while aligning the framework with the systematic reasoning of radiologists when they review a CXR. Our main contributions can be summarized as follows:

\begin{itemize}[noitemsep,nolistsep]
\item Our approach tackles the problem of radiology \emph{Findings} generation by treating it, in the first stage, as a multi-class detection problem, where bounding boxes are utilized to isolate the abnormalities. The entire image content is processed in a single forward pass, enabling the prediction of labeled bounding boxes with corresponding probabilities around the relevant abnormalities.
\item We harness the capabilities of LLMs to translate a list of abnormalities, along with their corresponding probabilities, into a conclusive \emph{Findings} report.

\end{itemize}
\begin{figure}[!b]
\centerline{\includegraphics[scale=0.2]{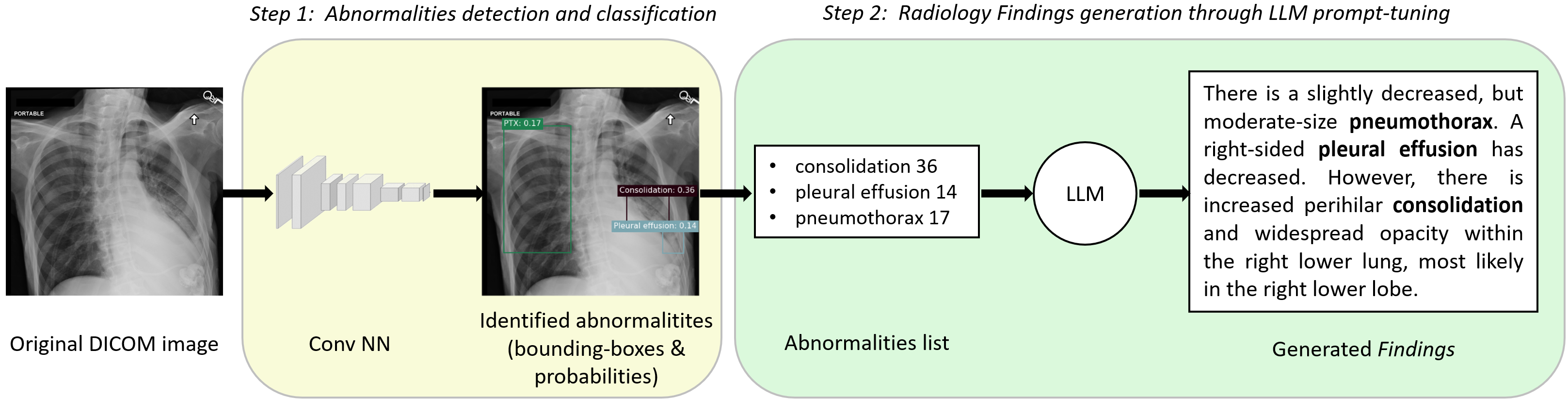}}
\label{fig:two step architecture}
\caption{Two-step architecture for radiology \emph{Findings} generation, starting from CXR images.}
\end{figure}

\section{Related Work}
Automated radiology report generation allows clinicians to interpret radiological images more easily than it would be possible with traditional anomaly detection and classification algorithms. Medical reports generation is a complex task, involving both the identification of relevant features in medical images (which may be affected by noise) and synthesizing this information into a coherent text that is easy  to understand. Hence, several previous works outlined the importance of fusing image and text features into a multimodal space \cite{chen2020uniter}, \cite{DBLP:journals/corr/abs-1908-02265}, \cite{DBLP:journals/corr/abs-2103-00020}. 

The generation of radiology reports necessitates cross-modal alignment of information from both medical images and textual reports. However, there are various challenges associated with this process, such as handling specialized vocabulary and domain knowledge, dealing with data heterogeneity, and having limited annotated training data available. Researchers have explored various approaches, including multimodal learning, natural language processing, and deep neural networks, to address the aforementioned challenges and improve the alignment process. The majority of these methods employ a conventional encoder-decoder architecture that relies on CNNs to encode the medical images \cite{vinyals2015show}, \cite{he2016deep}. The encoded information is then decoded into radiology reports using RNNs \cite{jing2017automatic} or Transformer networks \cite{chen2020generating}.


Some other approaches use contrastive learning \cite{dai2017contrastive} to capture important semantics in the images and learn meaningful representations \cite{endo2021retrieval}. This is because normal images and descriptions dominate the generated medical reports \cite{liu2021exploring}, as well as the number of normal images is higher than the abnormal ones \cite{shin2016learning}. These issues have been partially addressed through the use of contrastive attention mechanisms \cite{ma2021contrastive} or by maximizing the similarity between genuine image-text pairs and randomly paired ones \cite{zhang2022contrastive}.


When it comes to collaboration between models trained on different modalities, the image-to-text generation task is typically solved through feature-based methods. These methods embed both image and text features into a common feature space \cite{chen2020uniter} with the aim of minimizing distribution mismatch. However, our approach differs in that we sequentially transform features into final instances (image-to-abnormalities, abnormalities-to-report). We refer to this as collaborative knowledge training. By using separately trained models in a sequential prediction pipeline, we enable different neural architectures to fully utilize single-task specialization instead of relying on modality fusion.


The two natural language processing (NLP) tasks that receive the most attention in the context of radiology are Radiology Report Understanding and Radiology Summarization \cite{ghosh2023radling}. Radiology Report Understanding extracts information from a radiology report, including imaging study type and problem list, to aid in diagnosis and treatment planning. In contrast to the process of automatically creating radiology reports where specific \textit{Findings} are produced, radiology summarization involves creating a concise summary of a radiology report \textit{Findings}. Pre-trained or large language models are utilized in almost all recent works addressing these two NLP tasks \cite{karn2023shs}, \cite{ghosh2023radling}, \cite{DelbrouckRadSum23}.

\section{Dataset}
In the experiments conducted for this paper, we utilized a publicly available dataset called \emph{MIMIC-CXR} \cite{johnson2019mimic}. The dataset includes pairs of CXRs and their corresponding free-text medical reports, totaling 227,835 clinical studies from 65,379 patients. Each clinical study contains one or more images, typically a frontal or lateral view, and the dataset contains a total of 377,110 images.

\section{Methodology}


The proposed methodology for generating radiology reports follows a two-step approach (as shown in Figure 2). The first step involves detecting abnormalities in CXR images, treating this detection task as a multi-class problem. Each identified abnormality is localized within the input CXR as a bounding-box with an associated probability score. Our focus is on identifying multiple types of lung lesions, including nodules and masses, as well as pneumothorax.

In the second step, the methodology utilizes a LLM that has been fine-tuned to convert the list of abnormalities and their corresponding probabilities obtained in the first step into textual representations for the \emph{Findings} section of the radiology report. By leveraging the capabilities of the LLM, the methodology aims to generate comprehensive and accurate descriptions of the detected abnormalities in a format consistent with conventional radiology reporting.


Overall, this two-step methodology combines the power of computer vision techniques for abnormality detection in CXRs with the natural language generation capabilities of an LLM. This enables the automated generation of radiology reports that provide detailed \emph{Findings} based on the detected abnormalities and their respective probabilities.

\begin{figure}[!b]
\centerline{\includegraphics[scale=0.327]{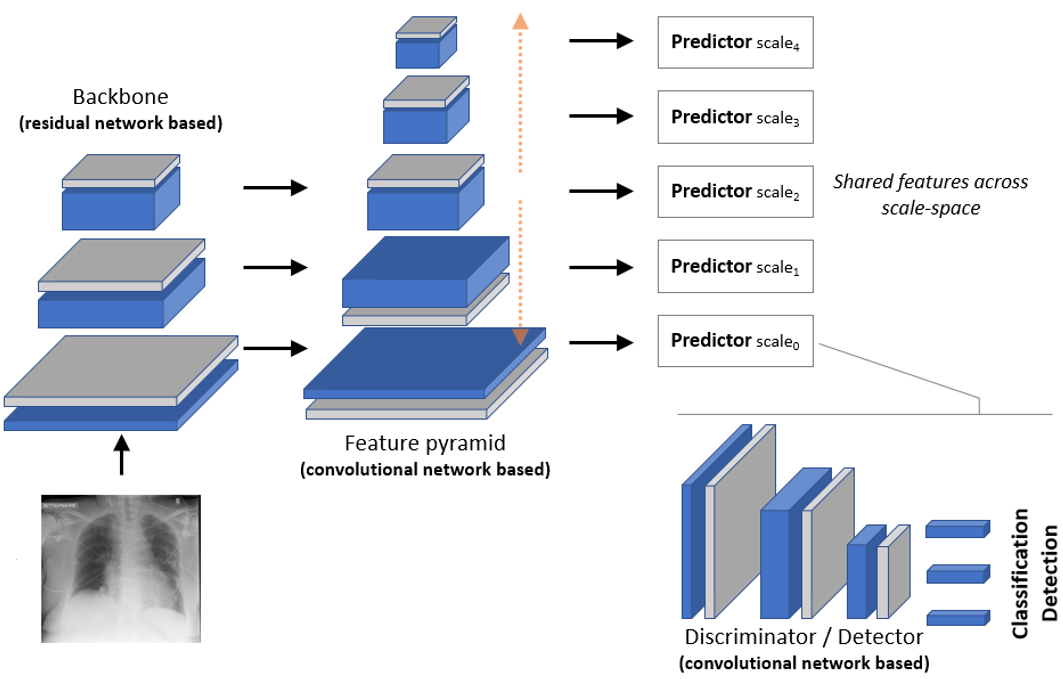}}
\label{fig:Architecture}
\caption{Architecture used for abnormalities detection and classification, as presented in \cite{ghesu2022self}.}
\end{figure}

\subsection{Abnormalities detection and classification}

Our main emphasis is on identifying lung abnormalities, such as lesions, consolidation, atelectasis, pleural effusion, pneumothorax, and others, from CXR images. The full list of abnormalities is presented in Table 1. These \emph{Findings} hold significant importance, as lung lesions can have long-term implications, such as pulmonary malignancy and cancer, while pneumothorax is an acute condition that can immediately endanger the patient's life. Pneumothoraces, often small in size (sometimes just a few millimeters in sectional width) or concealed by other structures like ribs, can be easily missed or overlooked due to their subtle nature.


For this task, we employed the multi-class detection system developed and described by Ghesu et al. in \cite{ghesu2022self} (Figure 3). The system uses a single learning model that combines classification and detection, allowing for the effective transfer of class-specific information during training. As shown in Figure 3, the network consists of an early feature extractor that generates candidate regions in an abstract feature space, followed by a discriminator sub-network that calculates probabilities for the presence of abnormalities in specific image regions of interest. By having a multi-scale structure, it can implicitly capture both global and local abnormalities in the image. Detection is done in an anchor box-free fashion, as proposed in \cite{tian2019fcos}, given the constraints in texture and scale, as well as the potential advantages of a multi-class training approach.



We chose to employ this system to conduct our experiments for multiple reasons:
\begin{itemize}[noitemsep,nolistsep]
\item The system, which is based on contrastive learning and online feature clustering, allows for self-supervised learning from multi-modality medical image data.
\item The system is the result of one of the largest machine learning experiments reported in the medical image data literature, in which large-scale self-supervised training experiments were conducted on datasets consisting of over 1,300,000 X-rays and over 105,000,000 multi-modality images.  
\item The system demonstrated significant performance improvements in the task of assessing abnormalities in CXR, including a notable increase in accuracy (averaging between 3-5\% in terms of AUC), improved robustness to various image augmentations such as intensity variations, rotations, or scaling, and accelerated model training convergence (over 50\% improvement).
\end{itemize}

\begin{table}[htbp]
\centering
\caption{List of \emph{Findings} reported in CXR images.}
\vspace*{2mm}
\label{tab:findings}
\begin{tabular} {cl}
\hline
\textbf{Class ID} & \textbf{Abnormality} \\
\hline
  0 & Background \\
  1 & Lesion \\
  2 & Consolidation \\
  3 & Device \\
  4 & Atelectasis \\
  5 & Pleural Effusion \\
  6 & Fibrosis \\ 
  7 & Pneumothorax (PTX) \\
  8 & Calcification \\ 
  9 & Fracture \\
  10 & Hilar Enlargement \\ 
  11 & Scoliosis \\
  12 & Eventration \\
  13 & Pneumoperitoneum \\ 
  14 & Hernia \\
  15 & Emphysema \\
  16 & Aortic Dilatation \\ 
  17 & Thickening \\
  18 & Tracheal Deviation \\ 
  19 & Subcutaneous Emphysema \\
  \hline
\end{tabular}
\end{table}

\subsection{Radiology Findings generation through LLM prompt-tuning}
Recent research has demonstrated that LLMs pre-trained on large amounts of textual data possess the ability to generalize relatively well to new tasks in a zero-shot setting. This means that LLMs can achieve reasonable performance on novel tasks, even if they have not been explicitly trained for those tasks \cite{brown2020language}. Examples include answering questions based on a given passage \cite{brown2020language}, \cite{radford2019language}, or generating summaries \cite{radford2019language}, \cite{duan2019zero}. 

Pretrained LLMs, however, face a couple of limitations in performing certain out-of-domain tasks \cite{bommasani2021opportunities}, especially in the medical field due to limited exposure to medical text data during the training phase. To tackle these limitations, a new approach has emerged known as multi-task prompted fine-tuning (MPF) or instruction tuning. This novel approach is a type of large-scale pretrain-and-prompt-tune paradigm where fine-tuning of large pretrained language models (PLMs) is carried out by using datasets that encompass diverse NLP tasks. The model is instructed which task to perform by using natural language prompts. Muennighoff et al. \cite{muennighoff2022crosslingual} implemented this method in their work, enabling their Bloomz LLM to effectively accomplish cross-lingual generalization in multilingual zero-shot instruction-based tasks.

In this work, we leverage the RadBloomz model \cite{karn2023shs}, which is a variant of the GPT-powered multi-task instruction-tuned Bloomz-7b1 \cite{muennighoff2022crosslingual} model. This model was adapted to the radiology domain by performing domain adapted pre-training (DAPT) on the \emph{MIMIC-IV} dataset \cite{johnson2023mimic}.


In this paper, we further fine-tune it to accomplish the task of generating radiology \emph{Findings} starting from CXR images. Following the RadBloomz approach, which uses \emph{Findings} as the prompt and employs cross-entropy loss to fine-tune Bloomz-7b1 by autoregressively comparing the generated summary with the ground-truth \emph{Impression}. We build our prompts by concatenating a list of abnormalities detected in CXR images and their corresponding probabilities into a text string, and we utilize the same cross-entropy loss to compare the generated \emph{Findings} with the corresponding ground-truth text. When creating the prompts, we list the detected abnormalities in a predefined order along with their corresponding probabilities. We chose to maintain a fixed order in the abnormalities list to reduce the complexity and confusion of the model, allowing the majority of the tuning to take place in the attention heads. 

If an abnormality is not detected (probability zero), we do not include it in the detection list. Additionally, if devices are detected within the input image, the \emph{'Device'} class is also not included in the abnormalities list. Before concatenating the prompt with the ground-truth \emph{Findings} section to create the input sequence, we append the ``TL;DR'' token at the end of the prompt. This is done to ensure consistency with the pretraining and instruction-tuning objectives of the base Bloom \cite{scao2022bloom} and intermediate Bloomz \cite{muennighoff2022crosslingual} models, as was done for RadBloomz.


To mitigate the risk of catastrophic forgetting in RadBloomz, the trainable parameters of Bloomz-7b1 undergo a reduction process in which all layers, except for the last one, are frozen. The ground-truth \emph{Findings} in the input sequence were obtained by removing sentences from the original \emph{Findings} in \emph{MIMIC-CXR} report that indicate the absence of certain abnormalities or refer to medical devices placed inside or on the surface of the body. We used Regular Expressions (Regex) for this filtering procedure.


\begin{figure}[t]
\centerline{\includegraphics[scale=0.205]{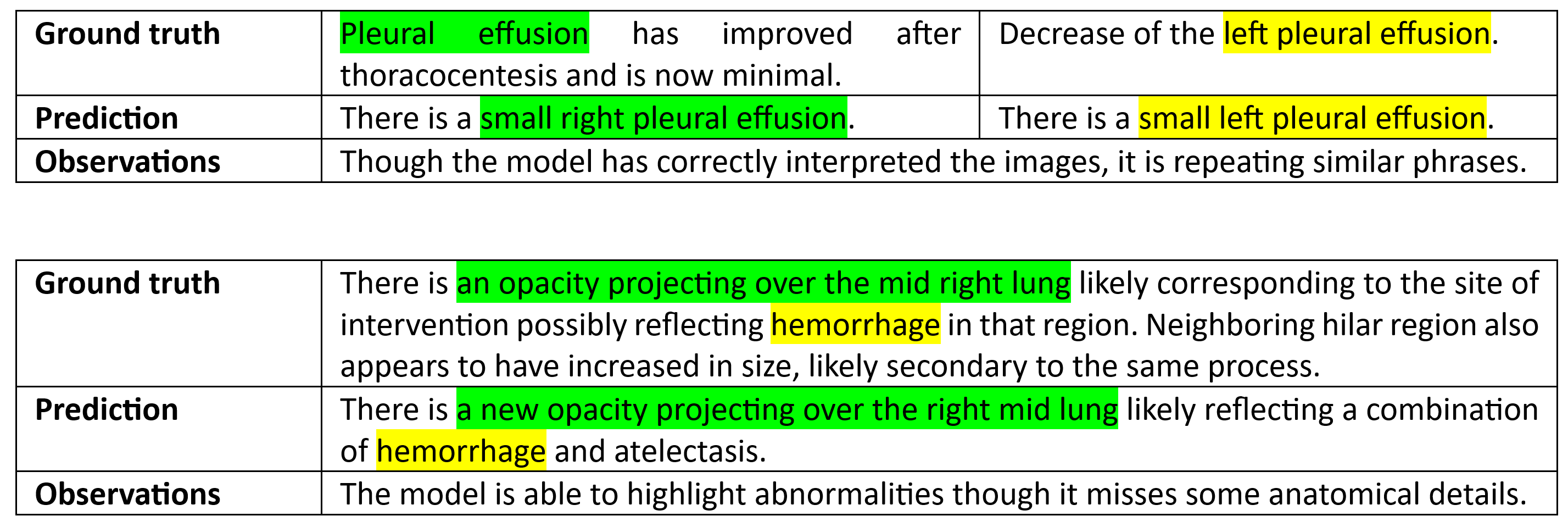}}
\label{fig:qualitative_example}
\caption{Qualitative analysis of generation accuracy depicting common shortcomings like repetition or model hallucination.}
\end{figure}

\section{Experiments and Results}
The experiments conducted in this paper aimed to fine-tune the radiology domain-adapted RadBloomz \cite{karn2023shs} LLM to generate radiology \emph{Findings} from a list of abnormalities detected in CXR images. The RadBloomz LLM was instructed to perform the \emph{Findings} generation task during the fine-tuning process by utilizing prompts, which guide the output generation process of the model and offer increased control (instruction-tuning). In this paper, the prompts are represented by lists of abnormalities that include both the names and probabilities of the abnormalities. These abnormalities were obtained by inferring on \emph{MIMIC-CXR} images using the multi-class detection system proposed in \cite{ghesu2022self}.


\subsection{LLM training setup}
We trained the LLM for 3 epochs using a batch size of 2 on 8 Nvidia Tesla A100 SXM4 GPUs (with 80 GB memory per GPU). To enhance efficiency, the system incorporates Deepspeed \cite{rasley2020deepspeed}, a deep learning optimization library that optimizes memory management and parallelism for faster training and inference times.
The data is split into train, validation, and test sets with a ratio of 70:10:20. As the average token length of \emph{Findings} is 128, this hyperparameter is fixed for the generation task.

\subsection{Experimental results}
In Table 2, we compare our method with state-of-the-art (SOTA) works, including conventional encoder-decoder architectures (CNN-RNN \cite{vinyals2015show} or CNN-Transformer \cite{liu2021exploring}), as well as methods that employ different learning paradigms, such as reinforcement learning \cite{qin2022reinforced}, curriculum learning \cite{liu2022competence}, or different approaches to learning multi-level cross-modal alignments \cite{yang2023unify}. According to the data in Table 2, our experimental results for \emph{Findings} generation on the \emph{MIMIC-CXR} dataset show a better ROUGE-L score than competing models, with an absolute improvement of 8.4 percentage points.


    \begin{table}[h]
      \centering
      \caption{Evaluation Metrics.}
      \vspace*{2mm}
      \label{tab:metrics}
      \begin{tabular}{ccccccc}
        \hline
        \textbf{Dataset} & \textbf{ROUGE-L} \\
        \hline
        ST \cite{vinyals2015show} & 0.263 \\
        CMCL \cite{liu2022competence} & 0.281 \\
        PPKED \cite{liu2021exploring} & 0.284 \\
        CMM+RL \cite{qin2022reinforced} & 0.287 \\
        UAR \cite{yang2023unify} & 0.289 \\
        OURS & \textbf{0.373} \\
        \hline
      \end{tabular}
    \end{table}

Hallucination and repetition are common issues encountered in generative models. As shown in Figure 4, our model demonstrates the ability to interpret images accurately, but there are instances where it fails to capture certain anatomical details. Moreover, the model tends to learn patterns for expressing various abnormalities, often generating such patterns with a higher probability. This behavior increases the risk of overlooking critical patient-specific details.

\subsection{Future work directions}

In this paper, we experimented with a fixed sequence length of 128 tokens for the generation process. Although the exact length of the \emph{Findings} report cannot be known beforehand, our future work will focus on dynamically predicting the possible report length to stop text generation, thus aiming to improve the model performance.

Another direction for future work involves incorporating the bounding-box coordinates of the detected abnormalities within a given CXR image during the prompt building process. This will enrich the generated sentences with valuable information about the localization of the identified abnormalities. Additionally, we can experiment with the original \emph{Findings} in \emph{MIMIC-CXR} without filtering out sentences that indicate the absence of certain abnormalities or refer to medical devices. This could also imply including in the abnormalities list the \emph{Findings} that were not detected (with probability zero).

Moreover, we aim to further improve the quality of the generated \emph{Findings} by exploring different prompts, such as "Generate the \emph{Findings} section for the following abnormalities, given their location and probabilities". This way, the system could more effectively guide the LLM to generate precise and informative reports aligned with the given input.

\subsection{Limitations}
There are a few limitations related to the training data we used, and we have listed them below:
\begin{itemize}[noitemsep,nolistsep]
    \item Our domain adaptation training data for RadBloomz was English reports only; therefore, it may not work out of the box in a multilingual setting.
    \item RadBloomz utilizes the \textit{MIMIC-IV} dataset for domain adaptation training, which might include overlapping data from \textit{MIMIC-CXR}. Thus, there is a potential risk of information leak in using RadBloomz.
\end{itemize}

\section{Conclusions}
In this study, we have demonstrated the significance of employing a two-step collaborative knowledge approach for generating radiology reports using abnormality detection in CXRs, followed by LLM generation of \emph{Findings}, as opposed to a direct image-to-text approach. Our \emph{Findings} highlight the potential of this approach in enhancing the efficiency and accuracy of automated report generation in radiology.


\section*{Acknowledgements}

The research leading to these results has received funding from the EEA Grants 2014–2021, under Project contract No. 33/2021. This work was supported by a grant from the Romanian National Authority for Scientific Research and Innovation, CCCDI–UEFISCDI, project number ERANET-PERMED-PROGRESS, within PNCDI III. This work received funding from the European Union’s Horizon Europe research and innovation programme under Grant Agreement No. 101057849 (DataTools4Heart project).


\bibliographystyle{elsarticle-num}
\bibliography{collaborativeVL}
\end{document}